\documentclass[10pt,twocolumn,letterpaper]{article}

\usepackage[pagenumbers]{cvpr} % To force page numbers, e.g. for an arXiv version

\newcommand{\TODO}[1]{\textbf{\color{red}[TODO: #1]}}
\renewcommand{\TODO}[1]{}

\usepackage[accsupp]{axessibility}

\definecolor{cvprblue}{rgb}{0.21,0.49,0.74}
\usepackage[pagebackref,breaklinks,colorlinks,allcolors=cvprblue]{hyperref}

\def\paperID{21846} % *** Enter the Paper ID here
\def\confName{CVPR}
\def\confYear{2026}

\title{Data Leakage Detection and De-duplication in Large Scale Geospatial Image Datasets}

\author{Yeshwanth Kumar Adimoolam$^{1,3}$\\
{\tt\small ya.kumar@edu.cut.ac.cy}
\and
Charalambos Poullis$^2$\\
{\tt\small charalambos@poullis.org}
\and
Melinos Averkiou$^1$\\
{\tt\small m.averkiou@cyens.org.cy}\\
\and 
    $^1$CYENS Centre of Excellence  \,\,\,\,\,\,\,\,
    $^2$ Concordia University  \,\,\,\,\,\,\,\,
    $^3$Cyprus University of Technology  \,\,\,\,\,\,\,\,
}

\begin{document}
\maketitle

\begin{abstract}
	In our study, we conducted a comprehensive analysis of three widely used datasets in the domain of building footprint extraction using deep neural networks: the INRIA Aerial Image Labelling dataset, SpaceNet 2: Building Detection v2, and the AICrowd Mapping Challenge datasets. Our experiments revealed several issues in the AICrowd Mapping Challenge dataset, where nearly 90\% (about 250k) of the training split images had identical copies, indicating a high level of duplicate data. Additionally, we found that approximately 56k of the 60k images in the validation split were also present in the training split, amounting to a 93\% data leakage.
  Furthermore, we present a data validation pipeline to address these issues of duplication and data leakage, which hinder the performance of models trained on such datasets. Employing perceptual hashing techniques, this pipeline is designed for efficient de-duplication and leakage identification. It aims to thoroughly evaluate the quality of datasets before their use, thereby ensuring the reliability and robustness of the trained models. Our code is available at \url{https://github.com/yeshwanth95/Hash_and_search}.
\end{abstract}

\section{Introduction}
\label{sec:intro}

In recent years, deep learning and pattern recognition techniques have had a significant impact on remote sensing.  In particular, a number of works have employed popular CNN architectures such as UNets and ResNets \citep{unet_paper, resnet_paper, ictnet, machine_learned_regPolyBuilding_Zorzi, Girard_2021_CVPR, Li_Zhao_Zhong_He_Lin_2021, Xu2022AccuratePM} as well as attention-based architectures \citep{polymapper, zorzi2022polyworld, Hu2022PolyBuildingPT} for tasks such as building footprint extraction, road network extraction, etc., which have important applications in downstream urban understanding tasks such as land use and land cover classification, urban planning, navigation, etc.
Deep learning solutions that can generalize to unseen data distributions require an abundance of data, which has a significant impact on their applicability. Hence, the availability of large-scale, high-resolution remote sensing image datasets is crucial for the success of such methods. In light of this, it is imperative to assess the quality of such datasets and their suitability for developing such deep-learning solutions.

Owing to the need for large-scale image datasets of high quality, the majority of deep learning literature tends to adopt widely used publicly available benchmark datasets to train and evaluate their methods and compare with existing state-of-the-art works. For building footprint extraction, the need for high-quality, curated datasets containing polygonal building footprints has prompted a significant amount of recent literature to utilize the AICrowd mapping challenge dataset \citep{mohanty2020deep} extensively for training and testing their methods, as well as for comparison with other state-of-the-art methods. Other popular datasets include the INRIA Aerial Image Labelling Dataset \citep{maggiori2017dataset} and the SpaceNet Building Detection dataset \citep{Etten2018SpaceNetAR}, however, these datasets either only provide raster building mask annotations or provide data in a non-standard format (e.g., GeoJSON, GeoTIFFs) for the computer vision and deep learning research community. 
The AICrowd dataset claims to solve this problem by providing large-scale, high-resolution satellite images with polygonal building footprint annotations made available in the popular MS-COCO format \citep{Lin2014MicrosoftCC}, allowing the immediate use of this dataset by the computer vision research community. Consequently, many recent works addressing the task of polygonal building footprint extraction have evaluated their methods using the AICrowd dataset, either in conjunction with other datasets or exclusively.

Due to the popularity of these datasets, we set out to investigate their quality and suitability for deep learning research. The INRIA Aerial Image Labelling and SpaceNet 2: Building Detection v2 datasets were observed to indicate no major issues during initial qualitative observations. However, the same cannot be said for the AICrowd Mapping Challenge dataset.
Due to its significant size and the availability of building footprint annotations, numerous state-of-the-art methods have employed the AICrowd dataset extensively for training and validation \citep{machine_learned_regPolyBuilding_Zorzi, Girard_2021_CVPR, Li_Zhao_Zhong_He_Lin_2021, Xu2022AccuratePM, polymapper, zorzi2022polyworld, Hu2022PolyBuildingPT}. However, a manual qualitative examination of this dataset reveals a plethora of issues. These include presence of exact and augmented duplicates within official splits of the dataset and data leakage across official splits. These issues have a considerable impact on the performance of downstream applications where this dataset is used for training and evaluating building footprint extraction methods.

This underlines the impetus for our study: the demand for an effective, easy-to-adopt pipeline to swiftly evaluate the quality of large-scale image datasets. Such pipelines could conserve the time and effort of the research community, allowing for more efficient use of available resources. Specifically, our contributions are as follows:

\begin{itemize}[]
	\item Our study presents a thorough analysis of three key large-scale remote-sensing datasets, with a particular focus on the AICrowd Mapping Challenge dataset. In this in-depth analysis, we identify and highlight critical issues such as extensive duplication, where nearly 89\% of the training images are duplicates (either exact or augmented), and significant data leakage, with about 97\% of the validation images also present in the training split.
	\item Complementing our analytical findings, we present a deduplication and leakage detection pipeline, specifically tailored for large-scale image datasets. By utilizing perceptual hashing methods to detect collisions, this pipeline is a practical and easy-to-adopt method for identifying and eliminating data duplication and leakage issues. Its application in analyzing the INRIA Aerial Image Labelling dataset, SpaceNet 2: Building Detection v2, and particularly the AICrowd Mapping Challenge, demonstrates its practicality and efficacy in enhancing dataset integrity, thereby contributing to improving the robustness of machine learning pipelines.
\end{itemize}

\section{Related Works}
\label{sec:related_works}

\textbf{State-of-the-art trained on the AICrowd Mapping Challenge dataset:}
Our experiments on the three datasets and subsequent analyses, as explained later in the paper, revealed significant issues with the AICrowd Mapping Challenge dataset. Furthermore, several recent studies \citep{polymapper, roof_extract_cgf, machine_learned_regPolyBuilding_Zorzi, weakSupervised_smallBuildings, Girard_2021_CVPR, Xu2022AccuratePM, Hu2022PolyBuildingPT, BuildMapper} that focus on the task of building footprint extraction from remotely sensed imagery have used this contaminated AICrowd Mapping Challenge dataset in their experiments to evaluate their proposed methods. In some studies \citep{zhao_buildingOutlineDelineation, autoBuildingExtract_ICG, zorzi2022polyworld}, this dataset has even been used \textit{exclusively} to benchmark and evaluate the proposed methods. Upon reviewing these works, it is evident that the AICrowd dataset has been extensively used in recent literature, which further motivates us to evaluate the quality of this dataset and inform the research community of the several issues discovered in this dataset.\\

\noindent
\textbf{Impact of Dataset Quality on Deep Learning Models:}
The issues discovered in the AICrowd dataset, and our subsequent analyses of methods using the dataset, make it clear that contamination in large image datasets negatively impacts the reusability, robustness, and generalization of models trained on such datasets. Excessive duplication and leakage in a benchmark dataset often lead to the trained models exhibiting overfitting behavior and performing poorly on out-of-distribution data at test time. Data leakage is prevalent in large-scale vision datasets, where overlap between training and evaluation data can inflate reported performance and undermine fair benchmarking; retrieval-based audits show that even subtle leakage can significantly affect downstream tasks \citep{pRamos_data_leakage}. Data duplication has scale-dependent effects, with larger models increasingly treating near-duplicates as exact memorization, which can harm generalization and highlights the need for rigorous deduplication in modern large-scale datasets \citep{kazdan2026scaledependentdataduplication}. Issues like data leakage also have implications for the fairness and reliability of machine learning benchmarks commonly used by the research community to evaluate ongoing research efforts. Therefore, such datasets must be carefully analyzed for issues such as data leakage, excessive duplication, etc., before being used for model training/evaluations. However, this has become increasingly difficult to perform for large image datasets such as the ImageNet dataset \citep{imagenet}, MS-COCO dataset \citep{Lin2014MicrosoftCC}, the Cityscapes datasets \citep{Cordts2016Cityscapes}, etc., which can have up to several million image-annotation pairs, and also newer datasets such as the LAION-5B dataset \citep{schuhmann2022laionb} that can even have billions of image-annotation pairs. Therefore, there is an imperative need to develop efficient methods to evaluate and mitigate dataset quality issues (such as duplication and data leakage) on such large image datasets.

\noindent
\textbf{De-duplication of Large Image Datasets:}
Recent research has focused on de-duplicating large image datasets using neural network feature representations of images to detect duplicates. In CE-Dedup \citep{ce_dedup}, the authors use a hashing-based image de-duplication technique to significantly reduce the size of the dataset while still maintaining the accuracy of downstream image classification tasks. In Jafari O. et al \citep{Jafari2021ASO}, the authors study the suitability of locality-based hashing in a variety of downstream applications, such as machine learning and image/video processing. The authors in \citep{QHash} present QHash, a hashing algorithm for image de-duplication in datasets containing images with small visual differences, such as medical images. While these hashing techniques may be suitable for detecting near duplicates in an image dataset, we only focus on detecting exact and augmented duplicates ($90^\circ$ rotations and flips) in our experiments for which perceptual hashing would be more suitable.

Furthermore, in contrast to the hashing-based approaches described above, more recent methods also employ \textit{self-supervised pretraining schemes} to learn image descriptors that are then used in identifying similar images in the dataset \citep{Pizzi_self_supervised_dedup, Zhang_unsupervised_dedup}. Although such pretraining approaches could potentially achieve a higher degree of de-duplication for specific datasets, pretraining on very large datasets can be challenging and may not generalize well to other substantially different datasets. In light of these recent works, we adopt a perceptual hashing strategy to investigate the degree of data duplication and leakage in the AICrowd dataset.
\section{Methodology}
\label{sec:methods}

\subsection{Perceptual Hashing for Duplicate Detection}

\begin{figure*}[!htbp]
	% \begin{center}
	% \end{center}
    \centering
    \includegraphics[width=0.8\textwidth]{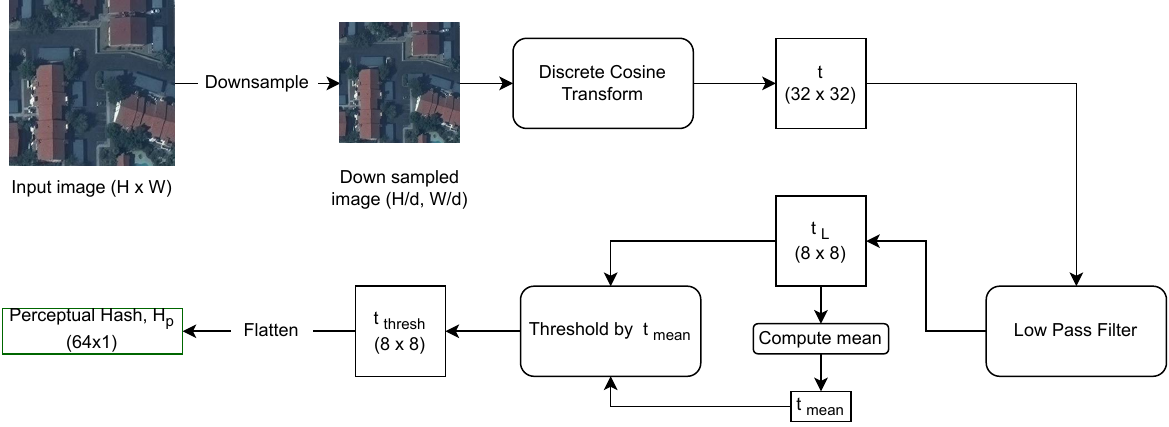}
	\caption{The pipeline used for computing the perceptual hash, $H_p$ of an image. The input image is first downsampled by a downsampling factor, $d$. The $32 \times 32$ discrete cosine transform, $t$ of the downsampled image is computed and the lowest frequencies, $t_L$ (top-left $8 \times 8$ values) are retained. Finally, $t_L$ is thresholded by the mean of the retained low frequencies and flattened to result in the $64-d$ perceptual hash, $H_p$ of the input image.}
	\label{fig:phash}
\end{figure*}

We present an effective method for detecting and eliminating data duplication and leakage in large-scale image datasets. The pipeline is independent of any particular dataset and it is based on the calculation of perceptual hashes of images in the dataset, as shown in Figure \ref{fig:phash}. This method efficiently identifies exact duplicates as well as augmented copies of images in a dataset. Augmented copies are images that have undergone transformations such as rotations and flips but remain inherently the same image.

We apply our method to the datasets described in Section \ref{sec:datasets_used}. The INRIA Aerial Image Labelling Dataset \citep{maggiori2017dataset} and the SpaceNet 2: Building Detection v2 dataset \citep{Etten2018SpaceNetAR} passed the scrutiny of our pipeline without revealing any major issues. These datasets consist of aerial and satellite images, respectively, with spatial resolutions of 0.3m, and are used in their officially provided train and test splits. However, when applying our pipeline to the AICrowd Mapping Challenge dataset, we discovered several issues as discussed in the next subsection.

\subsection{Data Leakage Removal \& De-duplication}
To illustrate the applicability of the proposed method and subsequent de-duplication, we use the AICrowd Mapping Challenge dataset as a representative example. Below we describe the steps taken to analyze and address the issues identified.

\noindent
\textbf{Initial Observations:} To determine the scope of data leakage between the official training and validation splits of the AICrowd dataset, we calculated the perceptual hashes, as shown in Figure \ref{fig:phash} of the images in the official training and validation splits and also their corresponding augmented versions. 
After computing the hashes for all images in the official splits provided in the AICrowd dataset, we checked for exact duplicates across the splits by searching for exact hash collisions. We observed significant data leakage between the official training and validation splits of the AICrowd dataset.

\noindent
\textbf{Leakage Detection in the Official Splits:} Further analysis showed that several additional images in the training split were augmented copies of images in the validation split. To detect such augmented duplicates, we augmented every image in the validation split with the following augmentation: 90$^\circ$, 180$^\circ$, 270$^\circ$ rotations, and horizontal and vertical flips. The perceptual hashes of this augmented validation set were then compared to that of the training images. In this case, we found that $38.72\%$ (108,707) of the official training images were exact or augmented duplicates of images found in the official validation split. Based on these findings, it is evident that a significant portion of the validation split appears multiple times in the training split of the AICrowd dataset, resulting in significant data leakage.

\noindent
\textbf{Eliminating Augmented Duplicates:}  The following procedure was adopted to address the issue of data leakage between the official training and validation splits of the AICrowd dataset. First, we augmented all images in the official train split with 90$^{\circ}$, 180$^{\circ}$, 270$^{\circ}$ rotations, and horizontal and vertical flips. Then we calculated the perceptual hashes of all images in the augmented train split, identified exact and augmented duplicates by detecting hash collisions, and retained only truly unique train images. The retained image from each set of duplicates was determined arbitrarily. We followed the same procedure for the official validation split to obtain a subset of unique images for final validation.

\noindent
\textbf{Removal of Data Leakage:} Finally, for these remaining images, we examined hash collisions between the train and validation splits and eliminated all instances of leaked validation images in the train split.
\section{Experiments and Results}
\label{sec:results}

\noindent
\textbf{Experimental details:} In our experiments, we used a perceptual hashing algorithm with a bit depth of 64 to compute and store the hashes of each image in a dataset. For hash collision detection, we treated exact hash matches as a collision, i.e., adopted a Hamming distance threshold of 0 between the computed hashes. Hash comparisons were made by a simple equality check, resulting in a highly efficient and fast-to-compute de-duplication/leakage detection workflow for large-scale image datasets. %All experiments were conducted on a machine with an AMD Epyc 7313 processor and 32GB of memory.
The hash computations and comparisons for all experiments were conducted on a machine with an AMD EPYC 7313 server-grade CPU after allocating 8 cores and 32GB of memory. The experiments revealed average runtimes of \textit{4ms per image} for hash computation and \textit{4ms per comparison} for hash comparisons.
	
	% The following average runtimes were observed in our experiments:
	% \begin{itemize}
		%     \item \textbf{Hash computation} = 4ms/image
		%     \item \textbf{Hash comparisons} = 4ms/comparison.
		% \end{itemize}

\subsection{Datasets}
\label{sec:datasets_used}

We have evaluated three popular benchmark datasets widely used for training deep neural networks on building and building footprint segmentation: INRIA Aerial Image Labelling dataset \citep{maggiori2017dataset}, SpaceNet 2: Building Detection v2 \citep{Etten2018SpaceNetAR}, and AICrowd Mapping Challenge \citep{mohanty2020deep}.

The \textbf{INRIA Aerial Image Labelling Dataset} \citep{maggiori2017dataset} consists of public domain aerial images and building footprint masks of size $5000\times 5000$ with a spatial resolution of 0.3m. The official train split consists of 180 such tiles with corresponding binary ground truth building masks. The official test split consists of another set of 180 images whose annotations are not publicly available. In our experiments, we split each image in the dataset into $250\text{px}\times250\text{px}$ non-overlapping patches to result in 72,000 patches in the train split and 72,000 in the test split. These splits were used in our de-duplication and leakage detection experiments.

The \textbf{SpaceNet 2: Building Detection v2 dataset} \citep{Etten2018SpaceNetAR} consists of 24,586 satellite image scenes across four areas (Las Vegas, Paris, Shanghai, and Khartoum). The images are of size $650\text{px}\times650\text{px}$ with a spatial resolution of 0.3m. The officially provided train and test splits were used in our de-duplication and leakage detection experiments.

The \textbf{AICrowd Mapping Challenge dataset} \citep{mohanty2020deep}, derived from the larger SpaceNet v2 challenge dataset, is composed of $300 \times 300$ RGB patches of WorldView 3 satellite images, each with a spatial resolution of 0.3m. The dataset is reasonably large, with 280,741 images in the training set and 60,317 images in the validation set. All images include MS-COCO annotations of polygonal building footprints \citep{Lin2014MicrosoftCC}.

\subsection{Results}

\noindent
\textbf{Evaluation of INRIA Aerial Image Labelling dataset: }
We used our de-duplication pipeline to evaluate the quality of the INRIA Aerial Image Labelling dataset \citep{maggiori2017dataset}. The results of these evaluations are presented in Table \ref{tab:quant_data_leakage_inria} from which it can be seen that there is negligible data leakage or duplication in the official splits of the INRIA Aerial Image Labelling dataset \citep{maggiori2017dataset}. 

\begin{table*}[!ht]
	\centering
	\resizebox{0.8\textwidth}{!}{\setlength{\tabcolsep}{1pt} \begin{tabular}{|l|l|c|c|c|}
			\hline
			Search Set & Targets Set & Total number of images in Search set & Number of targets in Search set & Overlap \(\%\) \\
			\hline\hline
			Official Train & Official Train & 72,000 & 2 & $2.78 \times 10^{-3}\%$\\
			& Official Test & 72,000 & 6 & $8.33 \times 10^{-3}\%$\\
			\hline
			Official Test & Official Train & 72,000 & 7 & $9.72 \times 10^{-3}\%$\\
			& Official Test & 72,000 & 4 & $5.56 \times 10^{-3}\%$\\
			\hline\hline
			Augmented Train & Augmented Train & 72,000 & 12 & $16.67 \times 10^{-3}\%$\\
			& Augmented Test & 432,000 & 58 & $13.42 \times 10^{-3}\%$\\
			\hline
			Augmented Test & Augmented Train & 432,000 & 86 & $19.91 \times 10^{-3}\%$\\
			& Augmented Test & 72,000 & 18 & $25 \times 10^{-3}\%$\\
			\hline
		\end{tabular}
	}
	\caption{\textbf{Data Leakage and Duplication.} Summary of the extent of data leakage/duplication in the train and test splits of the INRIA dataset. The `Official' train/test sets are those provided by the INRIA Aerial Image Labelling dataset \citep{maggiori2017dataset}. The `Augmented' train/test sets refer to those obtained after augmenting the official sets with 90$^\circ$, 180$^\circ$, 270$^\circ$ rotations, and horizontal and vertical flips.}
	\label{tab:quant_data_leakage_inria}
\end{table*}

In Figure \ref{fig:data_leakage_inria_supp} of the supplementary, we depict some qualitative examples of detected data leakage instances across the official train and test splits of the INRIA Aerial Image Labelling dataset \citep{maggiori2017dataset}. It is seen that the detected leaked samples were simply low-contrast images containing only water bodies or grasslands with little to no buildings. Therefore, these detected leaked samples could be treated as false positives, indicating that there is no real data leakage or duplication in the officially provided training and test splits. From Figure \ref{fig:data_leakage_inria_supp}, it can also be seen that the hashing technique detects leakage instances despite these patches being from different geographical locations. This is because the hashing technique is invariant to color and small structural changes. The technique can be made more sensitive to smaller structural changes by increasing the bit depth of the hashing algorithm, however, we found that a bit depth of 64 was sufficient for the scope of this study.

\noindent
\textbf{Evaluation of SpaceNet 2: Building Detection v2 dataset: }
The results of the duplication and data leakage evaluations conducted on the SpaceNet 2: Building Detection v2 dataset \citep{Etten2018SpaceNetAR} are presented in Table \ref{tab:quant_data_leakage_spacenet}. From Table \ref{tab:quant_data_leakage_spacenet}, it can be seen that the SpaceNet 2 dataset also exhibits negligible data leakage and duplication. 

\begin{table*}[!ht]
	\centering
	\resizebox{0.8\textwidth}{!}{\setlength{\tabcolsep}{1pt} \begin{tabular}{|l|l|c|c|c|}
			\hline
			Search Set & Targets Set & Total number of images in Search set & Number of targets in Search set & Overlap \(\%\) \\
			\hline\hline
			Official Train & Official Train & 10,593 & 105 & 0.991\%\\
			& Official Test & 10,593 & 96 & 0.906\%\\
			\hline
			Official Test & Official Train & 3,526 & 43 & 1.219\%\\
			& Official Test & 3,526 & 30 & 0.851\%\\
			\hline\hline
			Augmented Train & Augmented Train & 10,593 & 163 & 1.539\%\\
			& Augmented Test & 63,558 & 708 & 1.114\%\\
			\hline
			Augmented Test & Augmented Train & 21,156 & 313 & 1.479\%\\
			& Augmented Test & 3,526 & 43 & 1.219\%\\
			\hline
		\end{tabular}
	}
	\caption{\textbf{Data Leakage and Duplication.} Summary of the extent of data leakage/duplication in the train and test splits of the SpaceNet v2 dataset. The `Official' train/test sets are those provided by the SpaceNet v2 dataset \citep{Etten2018SpaceNetAR}. The `Augmented' train/test sets refer to those obtained after augmenting the official sets with 90$^\circ$, 180$^\circ$, 270$^\circ$ rotations, and horizontal and vertical flips.}
	\label{tab:quant_data_leakage_spacenet}
\end{table*}

In Figure \ref{fig:data_leakage_spacenet_supp} of the supplementary, we depict some qualitative examples of detected instances of leakage between the official train and test splits of the SpaceNet 2: Building Detection v2 dataset \citep{Etten2018SpaceNetAR}. In this case, as well, the detected leaked/duplicate samples were simply no data rasters, which are a common artifact of georeferenced satellite imagery. Therefore, these could also be considered false positives, indicating there is no real data duplication/leakage in the SpaceNet 2 dataset. Although there are very minor differences between the detected instances of leakage, the hashing algorithm is invariant to such minor structural differences.

\subsubsection{Evaluation of AICrowd Mapping Challenge dataset}

\begin{table*}[!ht]
	\centering
	\resizebox{0.8\textwidth}{!}{\setlength{\tabcolsep}{1pt} 
		
		\begin{tabular}{|l|l|c|c|c|}
			\hline
			Search Set & Targets Set & Total number of images in Search set & Number of targets in Search set & Overlap \(\%\) \\
			\hline\hline
			& Official Train & & 166,193 & 59.19\%\\
			Official Train & Official Val & 280,741 & 95,241 & 33.92\%\\
			& Official Test & & 95,884 & 34.15\%\\
			\hline
			& Official Train & & 56,368 & \textbf{93.45\%}\\
			Official Val & Official Val & 60,317 & 10,658 & 17.67\%\\
			& Official Test & & 20,642 & 34.22\%\\
			\hline
			& Official Train & & 56,608 & \textbf{93.26\%}\\
			Official Test & Official Val & 60,697 & 20,639 & 34.00\%\\
			& Official Test & & 10,701 & 17.63\%\\
			\hline\hline
			& Augmented Train & 280,741 & 251,403 & 89.55\%\\
			Augmented Train & Augmented Val & 1,684,446 & 642,741 & 38.16\%\\
			& Augmented Test & 1,684,446 & 645,205 & 38.30\%\\
			\hline
			& Augmented Train & 361,902 & 351,182 & \textbf{97.04\%}\\
			Augmented Val & Augmented Val & 60,317 & 46,151 & 76.51\%\\
			& Augmented Test & 361,902 & 139,330 & 38.49\%\\
			\hline
			& Augmented Train & 364,182 & 353,041 & \textbf{96.94\%}\\
			Augmented Test & Augmented Val & 364,182 & 139,349 & 38.26\%\\
			& Augmented Test & 60,697 & 46,483 & 76.58\%\\
			\hline
		\end{tabular}
	}
	\caption{\textbf{Data Leakage and Duplication.} Summary of the extent of data leakage/duplication in the train, validation, and test splits of the AICrowd dataset. The `Official' train/val/test sets are those provided by the AICrowd Mapping Challenge dataset \citep{mohanty2020deep}. The `Augmented' train/val/test sets refer to those obtained after augmenting the official sets with 90$^\circ$, 180$^\circ$, 270$^\circ$ rotations, and horizontal and vertical flips.}
	\label{tab:quant_data_leakage}
\end{table*}

% \begin{figure*}[!ht]
% 	\centering
% 	% \resizebox{\textwidth}{!}
% 	\includegraphics[width=0.7\textwidth]{figures/figure2_cropped.pdf}
% 	\caption{\textbf{Data Leakage.} Here we show examples of data leakage in the AICrowd dataset \citep{mohanty2020deep} (CC BY-NC-SA 4.0). We sample four images from the \textbf{validation split} and show duplicates occurring in the \textbf{training split}.}
% 	\label{fig:data_leakage}
% \end{figure*}

\begin{figure*}[!ht]
    % \widefigure
    \centering
    \resizebox{0.7\textwidth}{!}
    % \resizebox{\textwidth}{!}
    {%
    \setlength{\tabcolsep}{1.0pt} % default value: 6pt
\begin{tabular}{cccc}
    
    \textbf{Validation Img} & \textbf{Duplicate \#1} & \textbf{Duplicate \#2} & \textbf{Duplicate \#3} \\

    \includegraphics[width=0.2\textwidth]{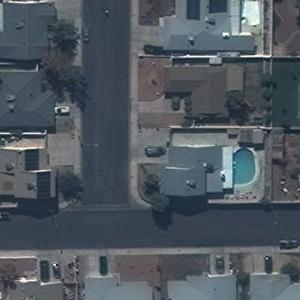}&
    \includegraphics[width=0.2\textwidth]{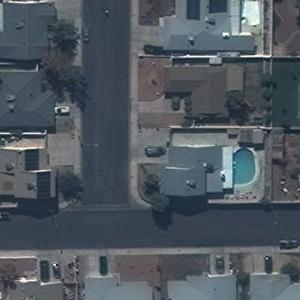}&
    \includegraphics[width=0.2\textwidth]{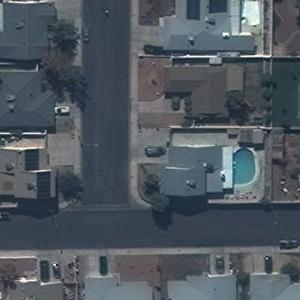}&
    \includegraphics[width=0.2\textwidth]{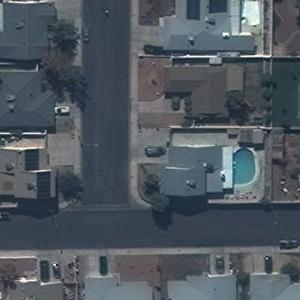}\\

    {\footnotesize 000000041307.jpg} & {\footnotesize 000000054203.jpg} & {\footnotesize 000000080396.jpg} & {\footnotesize 000000183508.jpg} \\

    \includegraphics[width=0.2\textwidth]{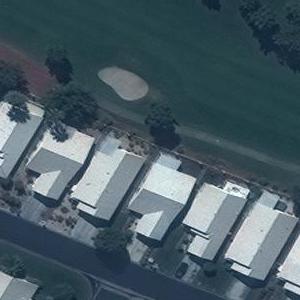}&
    \includegraphics[width=0.2\textwidth]{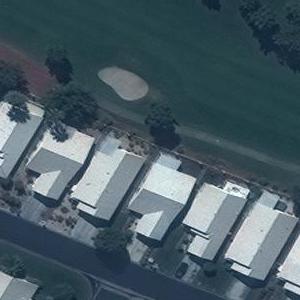}&
    \includegraphics[width=0.2\textwidth]{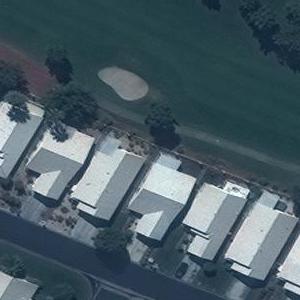}&
    \includegraphics[width=0.2\textwidth]{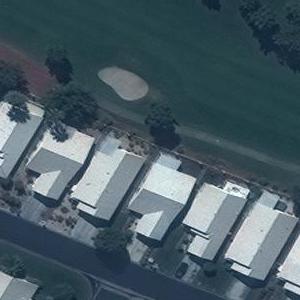}\\

    {\footnotesize 000000012224.jpg} & {\footnotesize 000000276960.jpg} & {\footnotesize 000000061321.jpg} & {\footnotesize 000000103221.jpg} \\

    \includegraphics[width=0.2\textwidth]{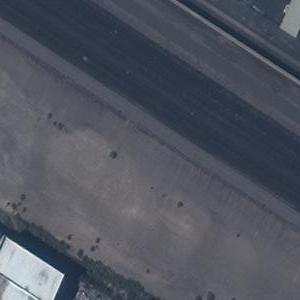}&
    \includegraphics[width=0.2\textwidth]{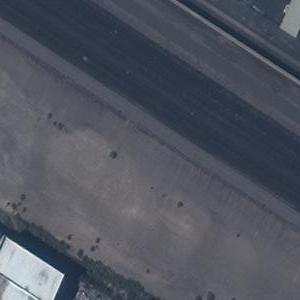}&
    \includegraphics[width=0.2\textwidth]{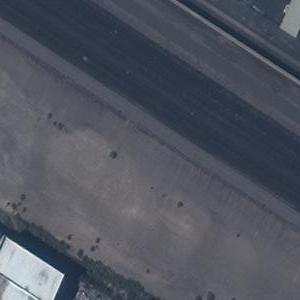}&
    \includegraphics[width=0.2\textwidth]{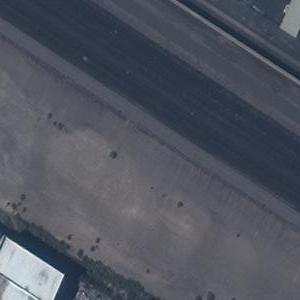}\\

    {\footnotesize 000000044948.jpg } & {\footnotesize 000000142052.jpg} & {\footnotesize 000000162593.jpg } & {\footnotesize 000000081730.jpg} \\

    \includegraphics[width=0.2\textwidth]{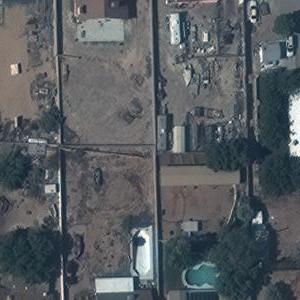}&
    \includegraphics[width=0.2\textwidth]{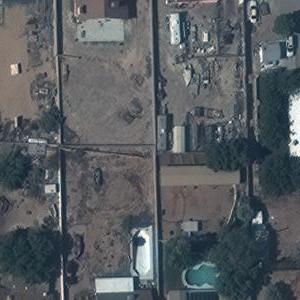}&
    \includegraphics[width=0.2\textwidth]{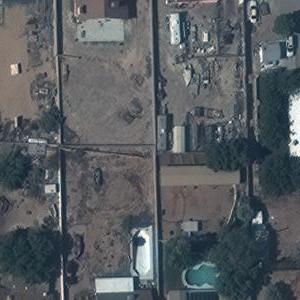}&
    \includegraphics[width=0.2\textwidth]{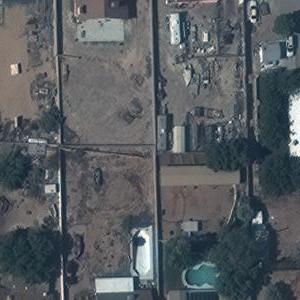}\\

    {\footnotesize 000000052295.jpg} & {\footnotesize 000000195238.jpg}  & {\footnotesize 000000217119.jpg} & {\footnotesize 000000112524.jpg} \\
    
\end{tabular}
}
    \caption{\textbf{Data Leakage.} Here we show examples of data leakage in the AICrowd dataset \cite{mohanty2020deep} (CC BY-NC-SA 4.0). We sample four images from the \textbf{validation split} and show duplicates occurring in the \textbf{training split}.}
    \label{fig:data_leakage}
\end{figure*}

\textbf{Data Leakage between train, validation, and test splits of the AICrowd dataset:}
Initial comparisons indicated that $93.45\%$ (56,368) of the 60,317 official validation images were also present in the training split. In contrast, $33.92\%$ (95,241) of the 280,741 official training images were exact duplicates of images in the validation split. Furthermore, $93.26\%$ (56,608) of the 60,697 official test split images were also present in the training split. The results of these experiments are presented in Table \ref{tab:quant_data_leakage} and some examples are illustrated in Figure \ref{fig:data_leakage} and Figures \ref{fig:data_leakage_crowdai_supp1} \& \ref{fig:data_leakage_crowdai_supp2} of the supplementary.

\noindent
\textbf{De-duplication of the AICrowd dataset:}
After removing duplicates and augmented duplicates from the official train and validation splits of the AICrowd dataset, the train split contained 29,338 unique images (out of the original 280,741) and the validation split contained 14,166 unique images (out of the original 60,317). From this subset, instances of leakage of validation images in the training split were identified and removed to further prune the training split to 15,392 images. This demonstrates that the AICrowd dataset exhibits severe redundancy and duplication of images.

\begin{figure*}[ht]
\centering
\resizebox{\textwidth}{!}{
\begin{tabular}{lcccccc}

\Huge PolyWorld \citep{zorzi2022polyworld} &
\includegraphics[width=0.6\linewidth]{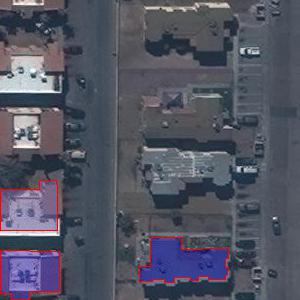} &
\includegraphics[width=0.6\linewidth]{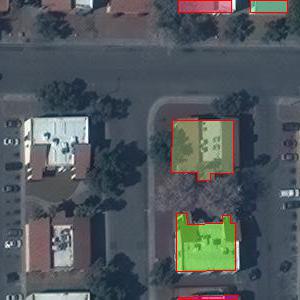} &
\includegraphics[width=0.6\linewidth]{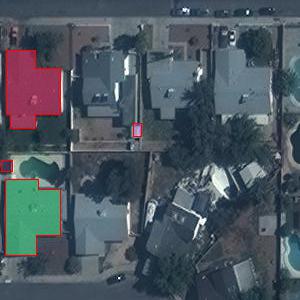} &
\includegraphics[width=0.6\linewidth]{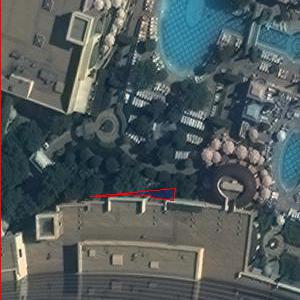} &
\includegraphics[width=0.6\linewidth]{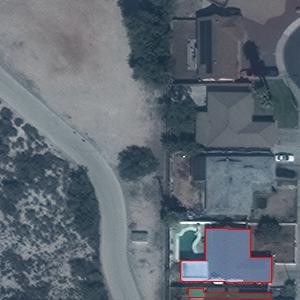} &
\includegraphics[width=0.6\linewidth]{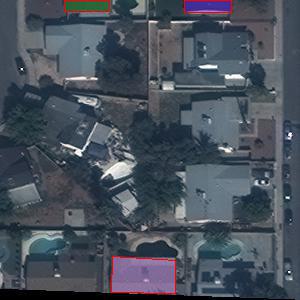} \\

\Huge HiSup \citep{Xu2022AccuratePM} &
\includegraphics[width=0.6\linewidth]{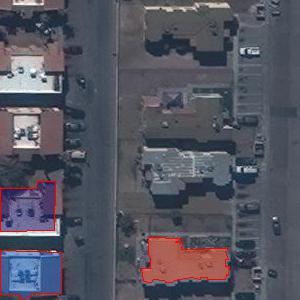} &
\includegraphics[width=0.6\linewidth]{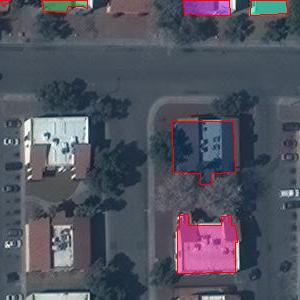} &
\includegraphics[width=0.6\linewidth]{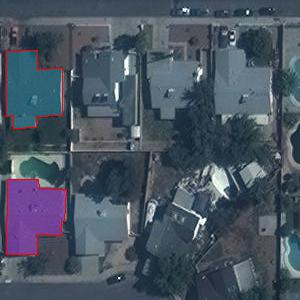} &
\includegraphics[width=0.6\linewidth]{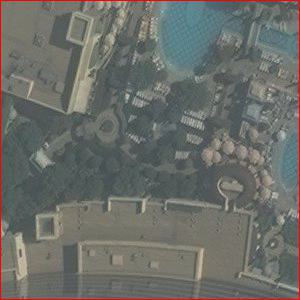} &
\includegraphics[width=0.6\linewidth]{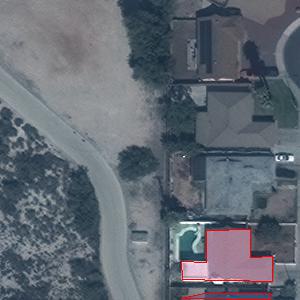} &
\includegraphics[width=0.6\linewidth]{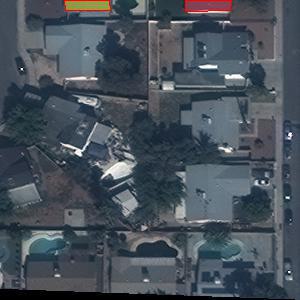} \\

\Huge Ground Truth &
\includegraphics[width=0.6\linewidth]{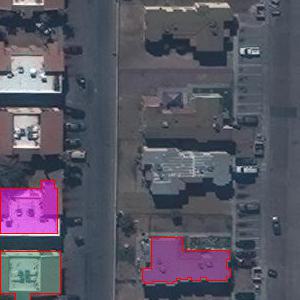} &
\includegraphics[width=0.6\linewidth]{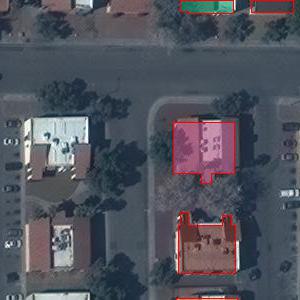} &
\includegraphics[width=0.6\linewidth]{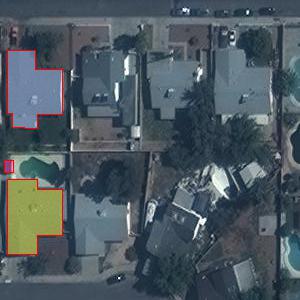} &
\includegraphics[width=0.6\linewidth]{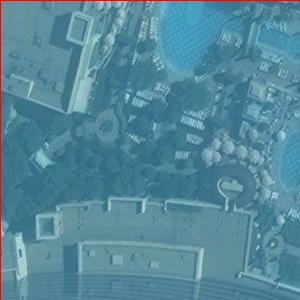} &
\includegraphics[width=0.6\linewidth]{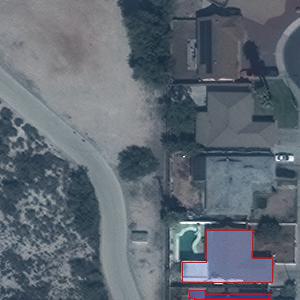} &
\includegraphics[width=0.6\linewidth]{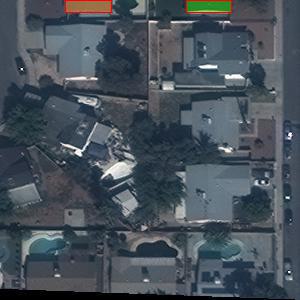} \\

\end{tabular}
}
\caption{
\textbf{Qualitative comparisons.} Examples from the original AICrowd \citep{mohanty2020deep} (CC BY-NC-SA 4.0) validation set where images are annotated incorrectly. We show example predictions from PolyWorld \citep{zorzi2022polyworld} (first row) and HiSup \citep{Xu2022AccuratePM} (second row). The ground truth is shown in the third row. In these examples, it can be seen that these methods replicate the incorrect/incomplete ground truth annotations, indicating overfitting due to data leakage between the train and validation splits.
}
\label{fig:wrongGT_comparison_crowdai}
\end{figure*}

\noindent
\textbf{Overfitting exhibited by methods reporting on the AICrowd dataset:}
Due to the presence of substantial duplication and data leakage in the official splits of the AICrowd dataset, it was discovered that several recently reported methods exhibit severe overfitting. This is particularly evident when these methods even replicate incorrect ground truth annotations from the training dataset. Qualitative examples of this behavior are shown in Figure \ref{fig:wrongGT_comparison_crowdai}. This explains why these methods achieve such high evaluation scores on the dataset.

% \begin{figure*}[!ht]
% 	\centering
% 	% \resizebox{\textwidth}{!}
% 	% {\setlength{\tabcolsep}{1.0pt} 
% 		\includegraphics[width=0.95\textwidth]{figures/figure3_cropped.pdf}
% 		\caption{\textbf{Qualitative comparisons.} Examples from the original AICrowd \citep{mohanty2020deep} (CC BY-NC-SA 4.0) validation set where images are annotated incorrectly. We show example predictions from PolyWorld \citep{zorzi2022polyworld} (first row) and HiSup \citep{Xu2022AccuratePM} (second row). The ground truth is shown in the third row. In these examples, it can be seen that these methods replicate the incorrect/incomplete ground truth annotations, indicating overfitting due to data leakage between the train and validation splits.}
% 		\label{fig:wrongGT_comparison_crowdai}
% \end{figure*}

\subsubsection{Comparison of the Perceptual Hashing Pipeline with Average Hashing}

\begin{table*}[!ht]
	\centering
	\resizebox{\textwidth}{!}{\setlength{\tabcolsep}{1pt} 
		
		\begin{tabular}{|l|l|c|c|c|}
			\hline
			Search Set & Targets Set & Total \# of images in Search set & Duplicates detected using PHash & Duplicates detected using AHash \\
			\hline\hline
			& Official Train & & 166,193 (59.2\%) & 167,829 (59.8\%)\\
			Official Train & Official Val & 280,741 & 95,241 (33.9\%) & 97,950 (34.9\%)\\
			& Official Test & & 95,884 (34.1\%) & 98,519 (35.1\%)\\
			\hline
			& Official Train & & 56,368 (93.4\%) & 56,431 (93.5\%)\\
			Official Val & Official Val & 60,317 & 10,658 (17.7\%) & 11,225 (18.6\%)\\
			& Official Test & & 20,642 (34.2\%) & 21,204 (35.1\%)\\
			\hline
			& Official Train & & 56,608 (93.3\%) & 56,740 (93.5\%)\\
			Official Test & Official Val & 60,697 & 20,639 (34.0\%) & 21,293 (35.1\%)\\
			& Official Test & & 10,701 (17.6\%) & 11,338 (18.7\%)\\
			\hline
		\end{tabular}
	}
	\caption{\textbf{Comparison of Duplicates and Leakage Detection Using Perceptual and Average Hashing Techniques.} Summary of the extent of data leakage/duplication in the official train, validation, and test splits of the AICrowd dataset \citep{mohanty2020deep}. The degree of duplication as a percentage of the various search sets are reported in the parentheses. It can be seen that the presence of data leakage and duplication in the AICrowd dataset is confirmed by both perceptual hashing (PHash) and average hashing (AHash) approaches.}
	\label{tab:quant_pipeline_comparisons}
\end{table*}

\begin{figure*}[!ht]
    % \widefigure
    \centering
    \resizebox{0.92\textwidth}{!}
    % \resizebox{\textwidth}{!}
    {%
    \setlength{\tabcolsep}{1.0pt} % default value: 6pt
\begin{tabular}{ccccc}
    
    \textbf{Hashing Algorithm} & \textbf{Test Img} & \textbf{Duplicate \#1} & \textbf{Duplicate \#2} & \textbf{Duplicate \#3} \\

    Perceptual Hashing & \includegraphics[width=0.2\textwidth]{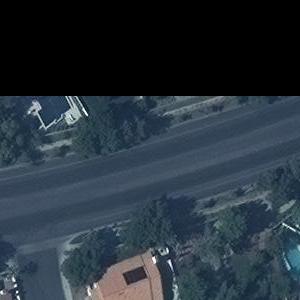}&
    \includegraphics[width=0.2\textwidth]{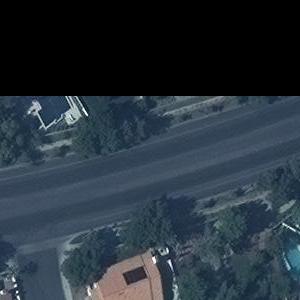}&
    \includegraphics[width=0.2\textwidth]{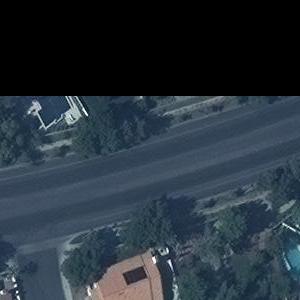}&
    \includegraphics[width=0.2\textwidth]{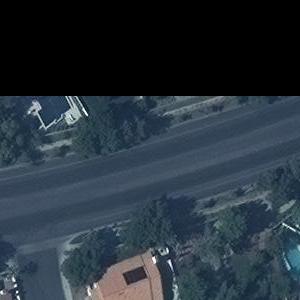}\\

     & {\footnotesize 000000000002.jpg} & {\footnotesize 000000261833.jpg} & {\footnotesize 000000119952.jpg} & {\footnotesize 000000198492.jpg} \\

     & \includegraphics[width=0.2\textwidth]{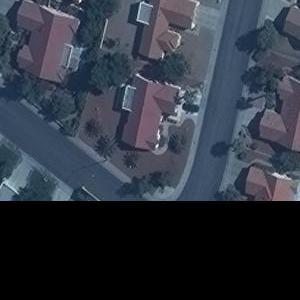}&
    \includegraphics[width=0.2\textwidth]{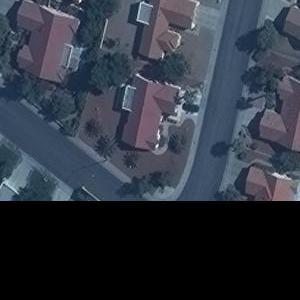}&
    \includegraphics[width=0.2\textwidth]{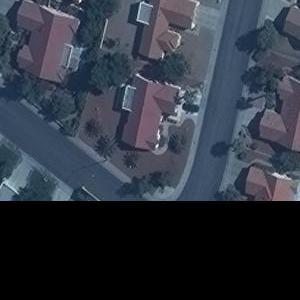}&
    \includegraphics[width=0.2\textwidth]{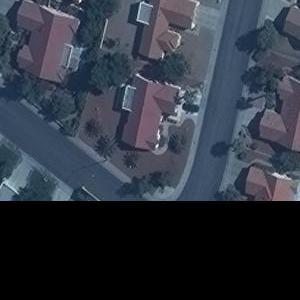}\\

     & {\footnotesize 000000050958.jpg} & {\footnotesize 000000185989.jpg} & {\footnotesize 000000271345.jpg} & {\footnotesize 000000125742.jpg} \\
     \hline \\

    Average Hashing & \includegraphics[width=0.2\textwidth]{figures/avg_hash_comparisons/000000000002.jpg}&
    \includegraphics[width=0.2\textwidth]{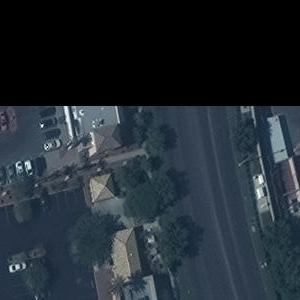}&
    \includegraphics[width=0.2\textwidth]{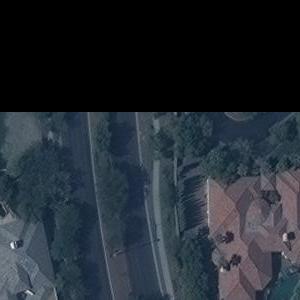}&
    \includegraphics[width=0.2\textwidth]{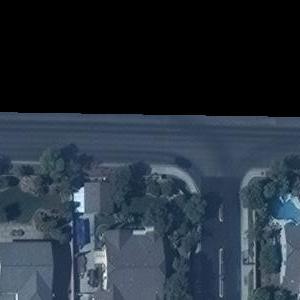}\\

     & {\footnotesize 000000000002.jpg} & {\footnotesize 000000249439.jpg} & {\footnotesize 000000245939.jpg} & {\footnotesize 000000011563.jpg} \\

     & \includegraphics[width=0.2\textwidth]{figures/avg_hash_comparisons/000000050958.jpg}&
    \includegraphics[width=0.2\textwidth]{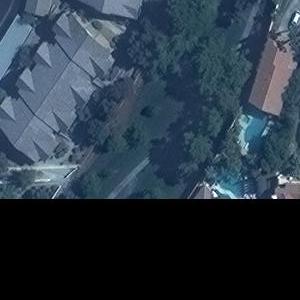}&
    \includegraphics[width=0.2\textwidth]{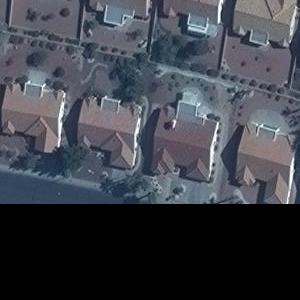}&
    \includegraphics[width=0.2\textwidth]{figures/avg_hash_comparisons/avg_hash/000000011284.jpg}\\

     & {\footnotesize 000000050958.jpg} & {\footnotesize 000000179970.jpg} & {\footnotesize 000000011284.jpg} & {\footnotesize 000000205692.jpg} \\

\end{tabular}
}
    \caption{\textbf{Qualitative Comparisons of Duplicates detected using Perceptual Hashing vs. Average Hashing.} Here we show examples of data leakage in the AICrowd dataset \cite{mohanty2020deep} (CC BY-NC-SA 4.0). We sample two images from the \textbf{test split} in the $1^{st}$ column and show duplicates occurring in the \textbf{training split} in the $2^{nd}$, $3^{rd}$, and $4^{th}$ columns. It can be seen that the Perceptual Hashing approach is less prone to false positives when compared to the Average Hashing approach.}
    \label{fig:average_hashing_false_positives}
\end{figure*}

To reasonably verify the results of the analyses conducted using the perceptual hashing pipeline, we also ran checks using a standard average hashing algorithm for detecting data leakage and duplicates across the official train, validation, and test splits of the AICrowd dataset \citep{mohanty2020deep}. The results of these comparisons are presented in Table \ref{tab:quant_pipeline_comparisons}. Therefore, it can be seen that analyses using both perceptual hashing and average hashing detect a similar extent of data leakage and duplication in the official train, validation, \& test splits of the AICrowd dataset. This confirms that the AICrowd dataset suffers from considerable data leakage and duplication issues. The minor difference in the detected number of duplicates is because average hashing is prone to false positives, where similar images are sometimes incorrectly flagged as duplicates, as shown in Figure \ref{fig:average_hashing_false_positives}. From these results, it is clear that perceptual hashing is less prone to false positive errors and hence is a more suitable choice for evaluating large-scale image datasets.

% \begin{figure*}[!ht]
% 	\centering
% 	% \resizebox{\textwidth}{!}
% 	\includegraphics[width=0.92\textwidth]{figures/figure8_cropped.pdf}
% 	\caption{\textbf{Qualitative Comparisons of Duplicates detected using Perceptual Hashing vs. Average Hashing.} Here we show examples of data leakage in the AICrowd dataset \citep{mohanty2020deep} (CC BY-NC-SA 4.0). We sample two images from the \textbf{test split} in the $1^{st}$ column and show duplicates occurring in the \textbf{training split} in the $2^{nd}$, $3^{rd}$, and $4^{th}$ columns. It can be seen that the Perceptual Hashing approach is less prone to false positives when compared to the Average Hashing approach.}
% 	\label{fig:average_hashing_false_positives}
% \end{figure*}

\section{Conclusion and Discussion}
\label{sec:discussion}

From the various quantitative and qualitative results presented, it can be observed that the deduplication pipeline is effective at detecting instances of duplication and leakage in large image datasets. The issues of leakage and duplication discovered in the AICrowd Mapping Challenge dataset render it unsuitable for benchmarking building footprint extraction methods without removing the leakage and duplication instances. This also potentially invalidates the quantitative metrics reported on this dataset by several preceding works. We also observe that the INRIA Aerial Image Labelling and SpaceNet 2: Building Detection v2 datasets are generally devoid of such major issues and could serve as more suitable datasets for benchmarking future research focusing on the task of building footprint detection.

\noindent
\textbf{Choice of Datasets: }The objective of this study was to evaluate the quality of the most common and popular geospatial datasets in the building footprint extraction literature. However, the proposed deduplication pipeline can be easily used for the assessment of any large-scale image dataset such as ImageNet \citep{imagenet}, VOC \citep{Everingham10VOC}, MS-COCO \citep{Lin2014MicrosoftCC}, Cityscapes \citep{Cordts2016Cityscapes}, etc.

\noindent
\textbf{Limitations:} Despite the effectiveness of the proposed pipeline, there are also some limitations worth noting. In the present pipeline, although the perceptual hashing algorithm is invariant to radiometric augmentations (such as brightness \& contrast changes), it is not inherently invariant to geometric augmentations such as rotation or flips. We overcome this limitation by augmenting the input images before the hash computation as part of the pipeline. The augmentations were chosen based on an initial visual inspection of the nature of duplications occurring in the datasets. Therefore, the choice of augmentations depends on the statistics of the dataset, i.e., some a priori information about the dataset is required before choosing appropriate augmentations. These limitations could be addressed in future research by developing more robust hashing algorithms that are inherently invariant to strong geometric \& radiometric transformations.

% \noindent
\section*{Acknowledgements}
This project has received funding from the EU H2020 Research and Innovation Programme and the Republic of Cyprus through the Deputy Ministry of Research, Innovation and Digital Policy (GA 739578). This project was also partly supported by the Natural Sciences and Engineering Research Council of Canada Grant RGPIN-2021-03479 (NSERC DG) and the MITACS Graduate Research Award IT34275.

{
    \small
    \bibliographystyle{ieeenat_fullname}
    \bibliography{main}

@String(CVPR= {IEEE Conf. Comput. Vis. Pattern Recog.})

@String(ICCV= {Int. Conf. Comput. Vis.})

@String(ICPR = {Int. Conf. Pattern Recog.})

@String(AAAI = {AAAI})

@String(CVPR  = {CVPR})

@String(ICCV  = {ICCV})

@String(ICPR  = {ICPR})

@inproceedings{unet_paper,
  author    = {Ronneberger, O. and Fischer, P. and Brox, T.},
  title     = {U-net: Convolutional networks for biomedical image segmentation},
  booktitle = {Medical Image Computing and Computer-Assisted Intervention -- MICCAI 2015},
  editor    = {Navab, N. and Hornegger, J. and Wells, W.~M. and Frangi, A.~F.},
  pages     = {234--241},
  publisher = {Springer International Publishing},
  address   = {Cham},
  year      = {2015}
}

@inproceedings{resnet_paper,
  author    = {He, K. and Zhang, X. and Ren, S. and Sun, J.},
  title     = {Deep residual learning for image recognition},
  booktitle = {2016 IEEE Conference on Computer Vision and Pattern Recognition (CVPR)},
  pages     = {770--778},
  year      = {2016},
  doi       = {10.1109/CVPR.2016.90}
}

@inproceedings{ictnet,
  author    = {Chatterjee, B. and Poullis, C.},
  title     = {On building classification from remote sensor imagery using deep neural networks and the relation between classification and reconstruction accuracy using border localization as proxy},
  booktitle = {2019 16th Conference on Computer and Robot Vision (CRV)},
  pages     = {41--48},
  year      = {2019},
  doi       = {10.1109/CRV.2019.00014}
}

@inproceedings{machine_learned_regPolyBuilding_Zorzi,
  author    = {Zorzi, S. and Bittner, K. and Fraundorfer, F.},
  title     = {Machine-learned regularization and polygonization of building segmentation masks},
  booktitle = {2020 25th International Conference on Pattern Recognition (ICPR)},
  pages     = {3098--3105},
  year      = {2021},
  doi       = {10.1109/ICPR48806.2021.9412866}
}

@inproceedings{Girard_2021_CVPR,
  author    = {Girard, N. and Smirnov, D. and Solomon, J. and Tarabalka, Y.},
  title     = {Polygonal building extraction by frame field learning},
  booktitle = {Proceedings of the IEEE/CVF Conference on Computer Vision and Pattern Recognition (CVPR)},
  pages     = {5891--5900},
  year      = {2021}
}

@inproceedings{Li_Zhao_Zhong_He_Lin_2021,
  author    = {Li, W. and Zhao, W. and Zhong, H. and He, C. and Lin, D.},
  title     = {Joint semantic-geometric learning for polygonal building segmentation},
  booktitle = {Proceedings of the AAAI Conference on Artificial Intelligence},
  volume    = {35},
  pages     = {1958--1965},
  year      = {2021}
}

@article{Xu2022AccuratePM,
  author  = {Xu, B. and Xu, J. and Xue, N. and Xia, G.-S.},
  title   = {Hisup: Accurate polygonal mapping of buildings in satellite imagery with hierarchical supervision},
  journal = {ISPRS Journal of Photogrammetry and Remote Sensing},
  volume  = {198},
  pages   = {284--296},
  year    = {2023},
  doi     = {10.1016/j.isprsjprs.2023.03.006}
}

@inproceedings{polymapper,
  author    = {Li, Z. and Wegner, J.~D. and Lucchi, A.},
  title     = {Topological map extraction from overhead images},
  booktitle = {2019 IEEE/CVF International Conference on Computer Vision (ICCV)},
  pages     = {1715--1724},
  year      = {2019},
  doi       = {10.1109/ICCV.2019.00180}
}

@inproceedings{zorzi2022polyworld,
  author    = {Zorzi, S. and Bazrafkan, S. and Habenschuss, S. and Fraundorfer, F.},
  title     = {Polyworld: Polygonal building extraction with graph neural networks in satellite images},
  booktitle = {Proceedings of the IEEE/CVF Conference on Computer Vision and Pattern Recognition},
  pages     = {1848--1857},
  year      = {2022}
}

@article{Hu2022PolyBuildingPT,
  author  = {Hu, Y. and Wang, Z. and Huang, Z. and Liu, Y.},
  title   = {Polybuilding: Polygon transformer for building extraction},
  journal = {ISPRS Journal of Photogrammetry and Remote Sensing},
  volume  = {199},
  pages   = {15--27},
  year    = {2023},
  doi     = {10.1016/j.isprsjprs.2023.03.021}
}

@article{mohanty2020deep,
  author  = {Mohanty, S.~P. and others},
  title   = {Deep learning for understanding satellite imagery: An experimental survey},
  journal = {Frontiers in Artificial Intelligence},
  volume  = {3},
  year    = {2020}
}

@inproceedings{maggiori2017dataset,
  author    = {Maggiori, E. and Tarabalka, Y. and Charpiat, G. and Alliez, P.},
  title     = {Can semantic labeling methods generalize to any city? The Inria aerial image labeling benchmark},
  booktitle = {IEEE International Geoscience and Remote Sensing Symposium (IGARSS)},
  year      = {2017},
  publisher = {IEEE}
}

@misc{Etten2018SpaceNetAR,
  author       = {Etten, A.~V. and Lindenbaum, D. and Bacastow, T.~M.},
  title        = {Spacenet: A remote sensing dataset and challenge series},
  howpublished = {ArXiv preprint arXiv:1807.01232},
  year         = {2018}
}

@inproceedings{Lin2014MicrosoftCC,
  author    = {Lin, T.-Y. and others},
  title     = {Microsoft COCO: Common objects in context},
  booktitle = {European Conference on Computer Vision},
  year      = {2014}
}

@article{roof_extract_cgf,
  author  = {Zhang, X. and Aliaga, D.},
  title   = {Procedural roof generation from a single satellite image},
  journal = {Computer Graphics Forum},
  volume  = {41},
  pages   = {249--260},
  year    = {2022},
  doi     = {10.1111/cgf.14472},
  note    = {Available at \url{https://onlinelibrary.wiley.com/doi/pdf/10.1111/cgf.14472}}
}

@inproceedings{weakSupervised_smallBuildings,
  author    = {Lee, J.-H. and Kim, C. and Sull, S.},
  title     = {Weakly supervised segmentation of small buildings with point labels},
  booktitle = {2021 IEEE/CVF International Conference on Computer Vision (ICCV)},
  pages     = {7386--7395},
  year      = {2021},
  doi       = {10.1109/ICCV48922.2021.00731}
}

@article{BuildMapper,
  author  = {Wei, S. and Zhang, T. and Ji, S. and Luo, M. and Gong, J.},
  title   = {Buildmapper: A fully learnable framework for vectorized building contour extraction},
  journal = {ISPRS Journal of Photogrammetry and Remote Sensing},
  volume  = {197},
  pages   = {87--104},
  year    = {2023},
  doi     = {10.1016/j.isprsjprs.2023.01.015}
}

@inproceedings{zhao_buildingOutlineDelineation,
  author    = {Zhao, W. and Persello, C. and Stein, A.},
  title     = {Building instance segmentation and boundary regularization from high-resolution remote sensing images},
  booktitle = {IGARSS 2020 - 2020 IEEE International Geoscience and Remote Sensing Symposium},
  pages     = {3916--3919},
  year      = {2020},
  doi       = {10.1109/IGARSS39084.2020.9324239}
}

@inproceedings{autoBuildingExtract_ICG,
  author    = {Wang, A. and Zhang, P.},
  title     = {Automatic building extraction based on boundary detection network in satellite images},
  booktitle = {2022 29th International Conference on Geoinformatics},
  pages     = {1--7},
  year      = {2022},
  doi       = {10.1109/Geoinformatics57846.2022.9963802}
}

@inproceedings{imagenet,
  author    = {Deng, J. and others},
  title     = {Imagenet: A large-scale hierarchical image database},
  booktitle = {2009 IEEE Conference on Computer Vision and Pattern Recognition},
  pages     = {248--255},
  year      = {2009},
  doi       = {10.1109/CVPR.2009.5206848}
}

@inproceedings{Cordts2016Cityscapes,
  author    = {Cordts, M. and others},
  title     = {The cityscapes dataset for semantic urban scene understanding},
  booktitle = {Proc. of the IEEE Conference on Computer Vision and Pattern Recognition (CVPR)},
  year      = {2016}
}

@inproceedings{schuhmann2022laionb,
  author    = {Schuhmann, C. and others},
  title     = {{LAION}-5b: An open large-scale dataset for training next generation image-text models},
  booktitle = {Thirty-sixth Conference on Neural Information Processing Systems Datasets and Benchmarks Track},
  year      = {2022}
}

@inproceedings{ce_dedup,
  author    = {Li, X. and Chang, L. and Liu, X.},
  title     = {Ce-dedup: Cost-effective convolutional neural nets training based on image deduplication},
  booktitle = {2021 IEEE Intl Conf on Parallel \& Distributed Processing with Applications, Big Data \& Cloud Computing, Sustainable Computing \& Communications, Social Computing \& Networking (ISPA/BDCloud/SocialCom/SustainCom)},
  pages     = {11--18},
  year      = {2021},
  doi       = {10.1109/ISPA-BDCloud-SocialCom-SustainCom52081.2021.00017}
}

@misc{Jafari2021ASO,
  author       = {Jafari, O. and Maurya, P. and Nagarkar, P. and Islam, K.~M. and Crushev, C.},
  title        = {A survey on locality sensitive hashing algorithms and their applications},
  howpublished = {ArXiv preprint arXiv:2102.08942},
  year         = {2021}
}

@inproceedings{QHash,
  author    = {Li, X. and Chang, L. and Liu, X.},
  title     = {Qhash: An efficient hashing algorithm for low-variance image deduplication},
  booktitle = {2021 IEEE 23rd Int Conf on High Performance Computing \& Communications; 7th Int Conf on Data Science \& Systems; 19th Int Conf on Smart City; 7th Int Conf on Dependability in Sensor, Cloud \& Big Data Systems \& Application (HPCC/DSS-SmartCity/DependSys)},
  pages     = {9--15},
  year      = {2021},
  doi       = {10.1109/HPCC-DSS-SmartCity-DependSys53884.2021.00029}
}

@inproceedings{Pizzi_self_supervised_dedup,
  author    = {Pizzi, E. and Roy, S.~D. and Ravindra, S.~N. and Goyal, P. and Douze, M.},
  title     = {A self-supervised descriptor for image copy detection},
  booktitle = {2022 IEEE/CVF Conference on Computer Vision and Pattern Recognition (CVPR)},
  pages     = {14512--14522},
  year      = {2022},
  doi       = {10.1109/CVPR52688.2022.01413}
}

@article{Zhang_unsupervised_dedup,
  author  = {Zhang, Z. and others},
  title   = {Dataset-driven unsupervised object discovery for region-based instance image retrieval},
  journal = {IEEE Transactions on Pattern Analysis and Machine Intelligence},
  volume  = {45},
  pages   = {247--263},
  year    = {2023},
  doi     = {10.1109/TPAMI.2022.3141433}
}

@article{Everingham10VOC,
  author  = {Everingham, M. and Van~Gool, L. and Williams, C.~K.~I. and Winn, J. and Zisserman, A.},
  title   = {The pascal visual object classes (voc) challenge},
  journal = {International Journal of Computer Vision},
  volume  = {88},
  pages   = {303--338},
  year    = {2010}
}

@INPROCEEDINGS{pRamos_data_leakage,
  author={Ramos, Patrick and Ramos, Ryan and Garcia, Noa},
  booktitle={2025 IEEE/CVF International Conference on Computer Vision Workshops (ICCVW)}, 
  title={Data Leakage in Visual Datasets}, 
  year={2025},
  volume={},
  number={},
  pages={6368-6378},
  doi={10.1109/ICCVW69036.2025.00661}}

@misc{kazdan2026scaledependentdataduplication,
      title={Scale Dependent Data Duplication}, 
      author={Joshua Kazdan and Noam Levi and Rylan Schaeffer and Jessica Chudnovsky and Abhay Puri and Bo He and Mehmet Donmez and Sanmi Koyejo and David Donoho},
      year={2026},
      eprint={2603.06603},
      archivePrefix={arXiv},
      primaryClass={cs.LG},
      url={https://arxiv.org/abs/2603.06603}, 
}
}

\clearpage
\setcounter{page}{1}
\maketitlesupplementary

\section{Data Leakage and Overfitting}
To quantify the effect of train–test leakage, we partition the test set into seen images (exact duplicates of training images) and unseen images (no overlap with training). Using the AICrowd pretrained checkpoint of HiSup \cite{Xu2022AccuratePM}, we observe a pronounced discrepancy in polygonal segmentation performance between these two subsets, shown in Table \ref{tab:leakage_hisup}. Performance on seen test images is much higher than on unseen ones, indicating that the model primarily memorizes seen samples rather than generalizing to novel images. As a result, the originally reported test performance of HiSup \cite{Xu2022AccuratePM} is inflated, demonstrating overfitting induced by data leakage.

\begin{table}[h]
\centering
% \small
\caption{Polygonal segmentation results of HiSup \cite{Xu2022AccuratePM} on the AICrowd test set, split into seen \& unseen images.}
\label{tab:leakage_hisup}
\resizebox{0.7\linewidth}{!}{
\begin{tabular}{lcccccc}
\toprule
\textbf{Subset} &
\textbf{AP} &
\textbf{AP$_{50}$} &
\textbf{AP$_{75}$} &
\textbf{AR} \\
\midrule
All     & 79.4 & 92.7 & 85.3 & 81.5 \\
Seen    & 79.7 & 92.7 & 85.4 & 81.9 \\
Unseen  & 65.4 & 87.1 & 74.5 & 69.3 \\
\bottomrule
\end{tabular}}
\end{table}

\section{PHash vs AHash}
To evaluate the effect of the hashing algorithm on the duplicate detection pipeline, we constructed a benchmark dataset of 10,000 images with known ground truth duplicates. The source images were drawn from a unique subset of the AICrowd validation set, ensuring no pre-existing duplicates. The dataset comprises 2,501 duplicate groups (7,501 images) \& 2,499 purely unique images, yielding a 50\% duplicate ratio. Each duplicate group contains one original image and two augmented variants, for a total of 5,000 intentionally created duplicates. The augmentations were sampled uniformly from six transformation types: exact copies (781), rotations of 90° (841), 180° (861), 270° (856), horizontal flips (802), \& vertical flips (859). As shown in Table \ref{tab:hash_comparison}, perceptual hashing (pHash) achieved near-perfect performance while average hashing (aHash) showed lower performance. This shows that pHash is more robust to geometric transformations and less prone to false positives on this dataset.

\begin{table}[h]
\centering
\caption{Duplicate detection on the benchmark dataset.}
\label{tab:hash_comparison}
\resizebox{0.9\linewidth}{!}{
\begin{tabular}{lccccc}
\toprule
\textbf{Method} & \textbf{Precision} & \textbf{Recall} & \textbf{F1 Score} & \textbf{FP} & \textbf{FN} \\
\midrule
pHash  & 1.0000    & 0.9997 & 0.9999   & 0               & 2               \\
aHash  & 0.9164    & 0.9709 & 0.9429   & 664             & 218             \\
\bottomrule
\end{tabular}}
\end{table}

\section{Additional Qualitative Examples}

\begin{figure*}[ht]
	\centering
	% \resizebox{\textwidth}{!}
	\includegraphics[width=\textwidth]{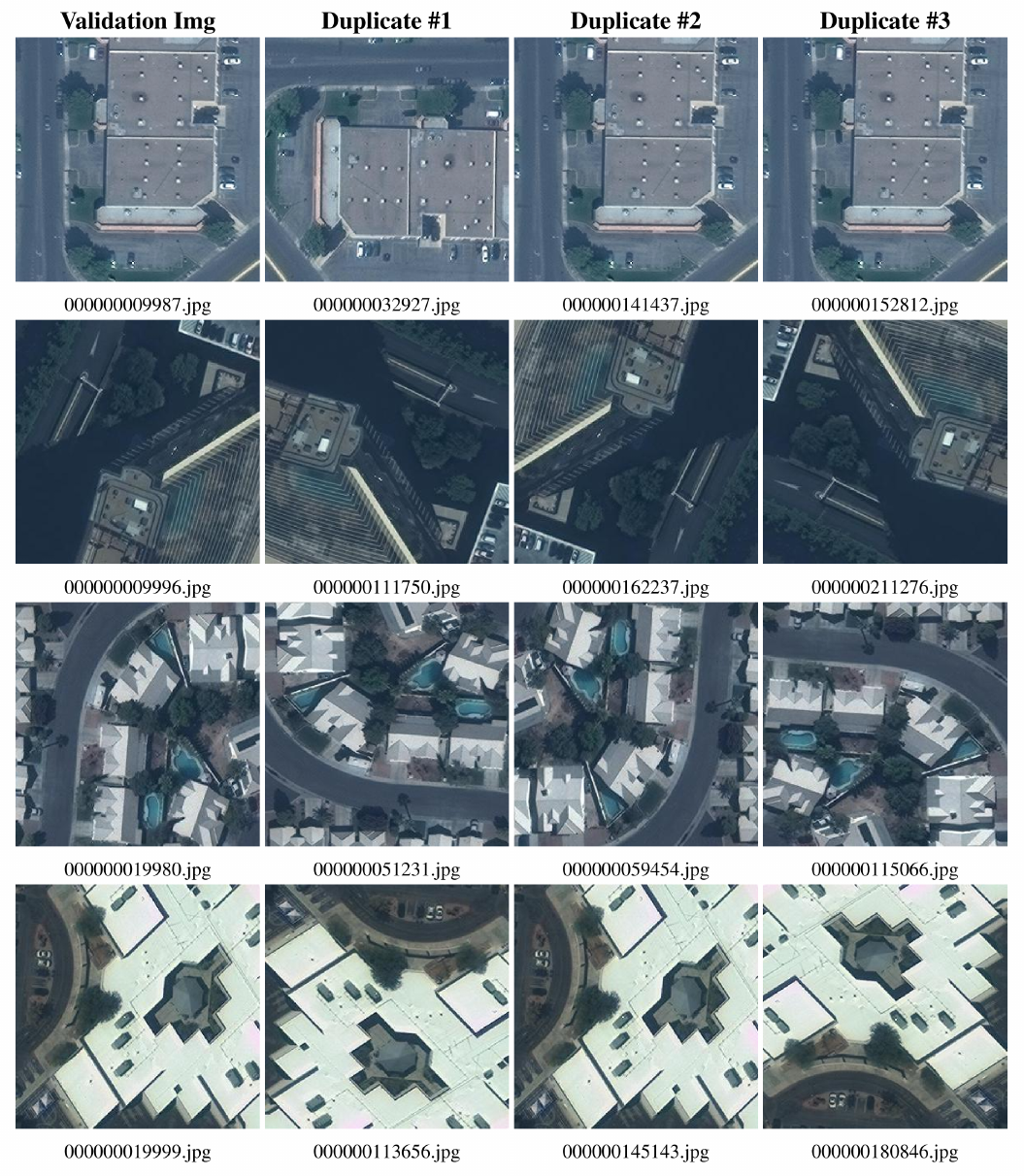}
	\caption{\textbf{Additional examples of data leakage.} Here we show additional examples of data leakage in the AICrowd Mapping Challenge dataset \citep{mohanty2020deep} (CC BY-NC-SA 4.0). We sample four images from the \textbf{validation split} in column 1 and show duplicates occurring in the \textbf{training split} in columns 2, 3, and 4.}
	\label{fig:data_leakage_crowdai_supp1}
\end{figure*}

\begin{figure*}[ht]
	\centering
	% \resizebox{\textwidth}{!}
	\includegraphics[width=\textwidth]{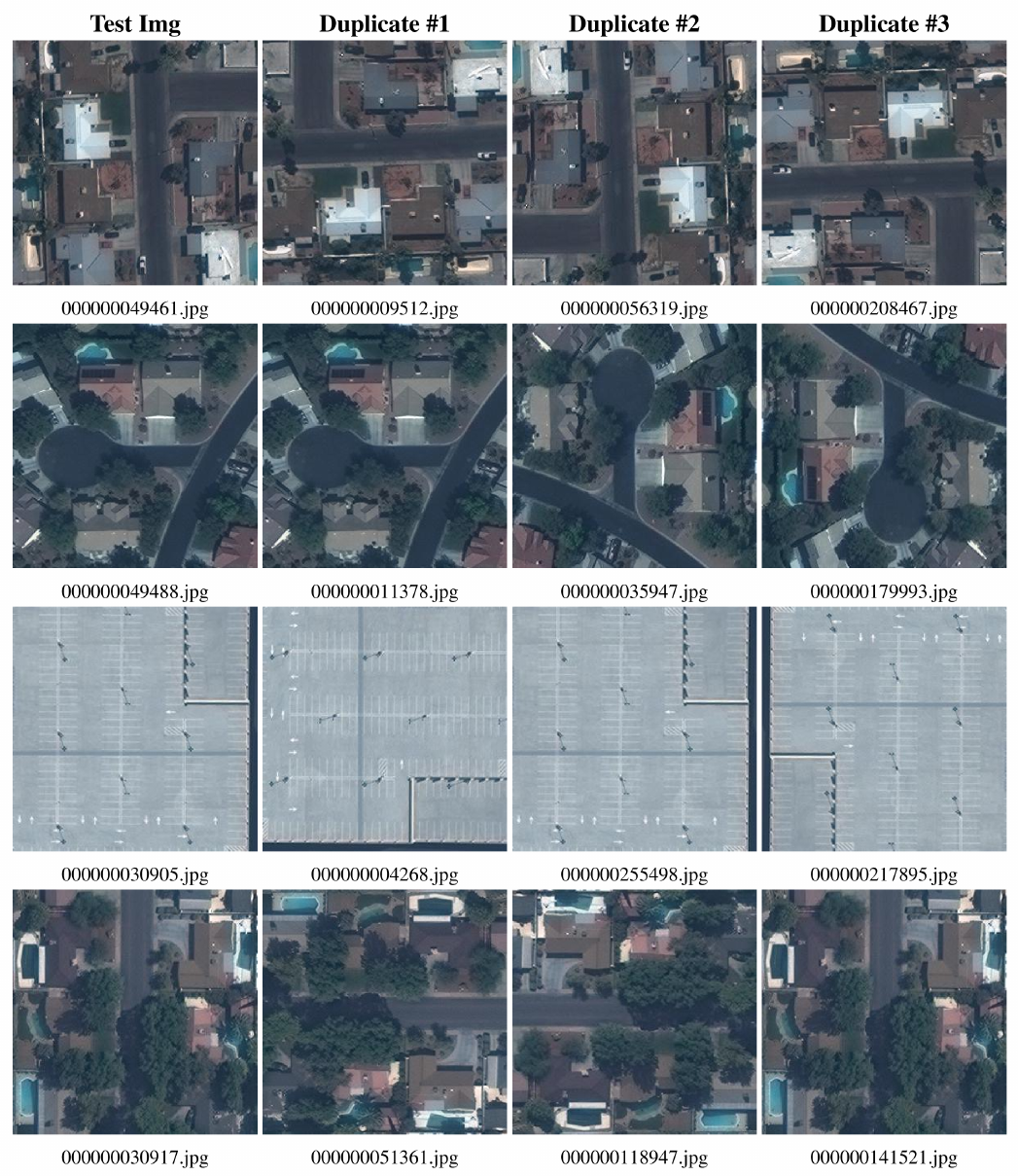}
	\caption{\textbf{Additional examples of data leakage.} Here we show additional examples of data leakage in the AICrowd Mapping Challenge dataset \citep{mohanty2020deep} (CC BY-NC-SA 4.0). We sample four images from the \textbf{test split} in column 1 and show duplicates occurring in the \textbf{training split} in columns 2, 3, and 4.}
	\label{fig:data_leakage_crowdai_supp2}
\end{figure*}

\begin{figure*}[!ht]
	\centering
	% \resizebox{0.67\textwidth}{!}
	\includegraphics[width=0.65\textwidth]{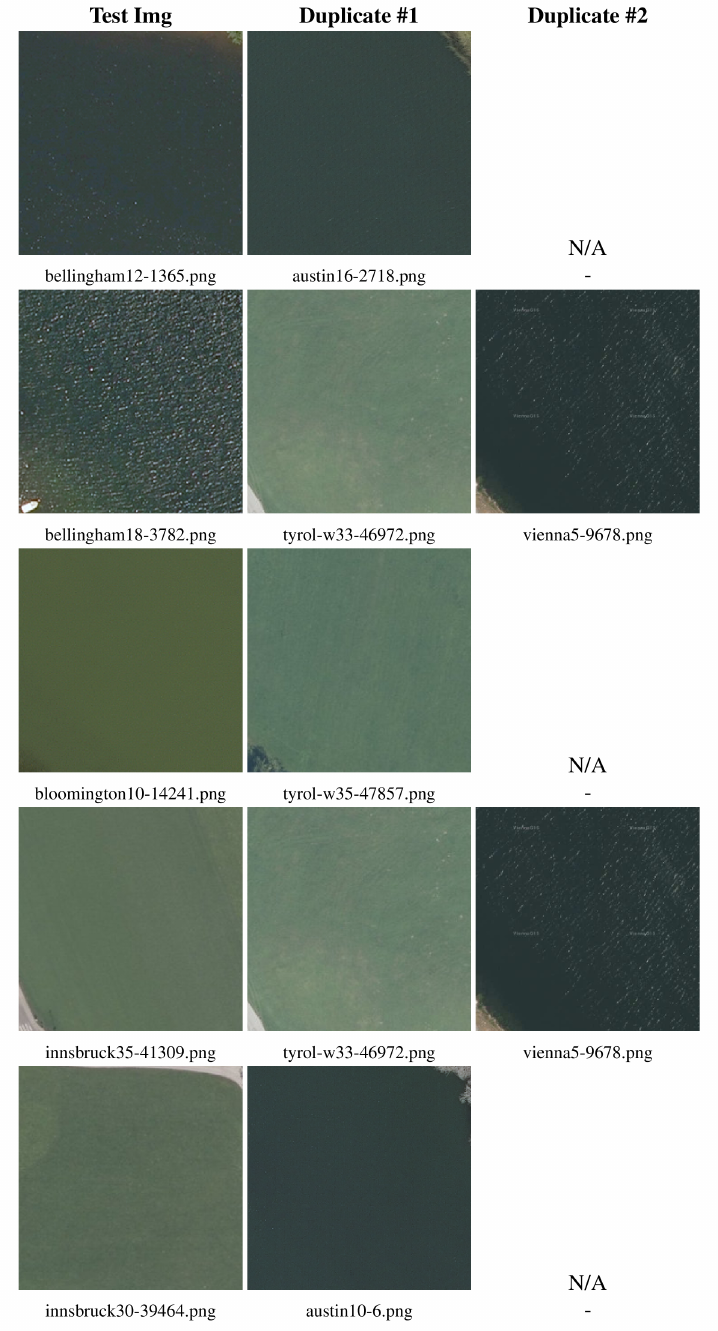}
	\caption{\textbf{False positive examples of data leakage.} Here we show falsely detected examples of data leakage in the INRIA Aerial Image Labelling dataset \citep{maggiori2017dataset}. We sample images from the \textbf{test split} in column 1 and show duplicates occurring in the \textbf{training split} in columns 2 and 3.}
	\label{fig:data_leakage_inria_supp}
\end{figure*}

\begin{figure*}[!ht]
	\centering
	% \resizebox{\textwidth}{!}
	\includegraphics[width=\textwidth]{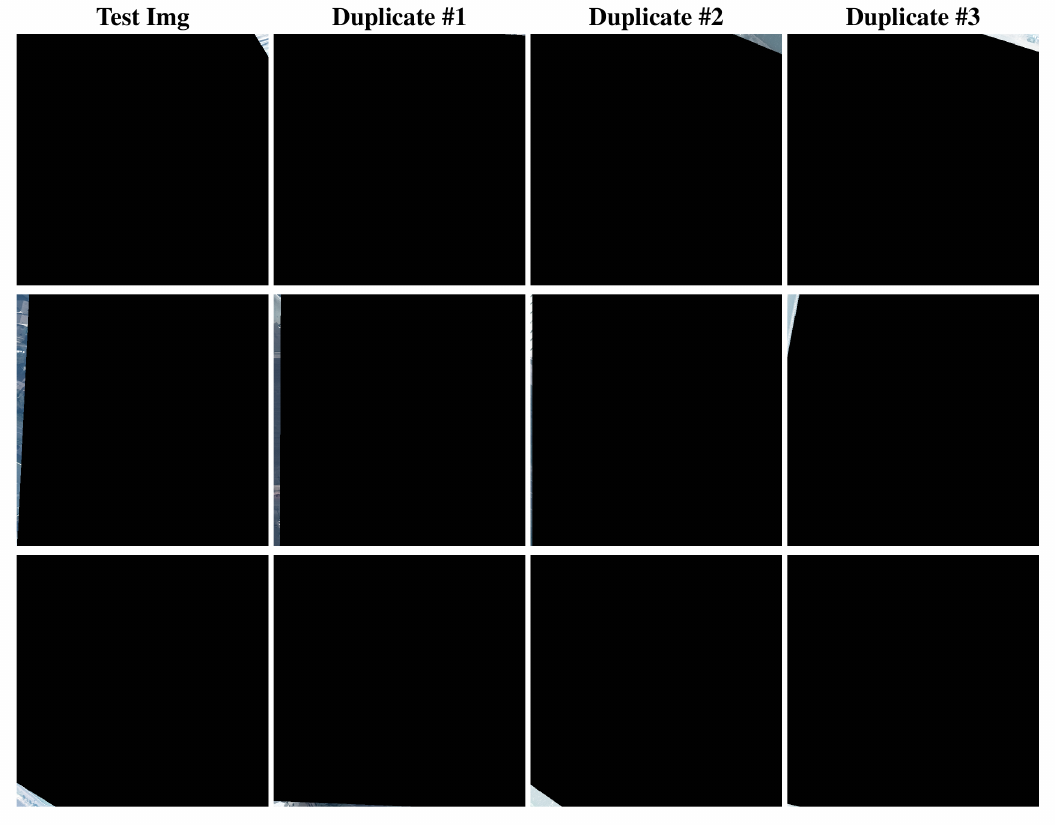}
	\caption{\textbf{False positive examples of data leakage.} Here we show examples of falsely detected examples of data leakage in the SpaceNet 2: Building Detection v2 dataset \citep{Etten2018SpaceNetAR} (CC BY-SA 4.0). We sample four images from the \textbf{test split} in column 1 and show duplicates occurring in the \textbf{training split} in columns 2, 3, and 4.}
	\label{fig:data_leakage_spacenet_supp}
\end{figure*}

In this supplementary material, we show qualitative examples of data leakage and duplication discovered in the AICrowd Mapping Challenge dataset \citep{mohanty2020deep} in Figures \ref{fig:data_leakage_crowdai_supp1} and \ref{fig:data_leakage_crowdai_supp2}. Additionally, in the case of the INRIA Aerial Image Labelling Dataset \citep{maggiori2017dataset} and the SpaceNet 2 Building Detection v2 dataset \citep{Etten2018SpaceNetAR}, we also depict some examples of the false positive duplicates identified by the deduplication pipeline in Figures \ref{fig:data_leakage_inria_supp} and \ref{fig:data_leakage_spacenet_supp} respectively.

%\section{False Positives}

% \section{Rationale}
% \label{sec:rationale}
% % 
% Having the supplementary compiled together with the main paper means that:
% % 
% \begin{itemize}
% \item The supplementary can back-reference sections of the main paper, for example, we can refer to \cref{sec:intro};
% \item The main paper can forward reference sub-sections within the supplementary explicitly (e.g. referring to a particular experiment); 
% \item When submitted to arXiv, the supplementary will already included at the end of the paper.
% \end{itemize}
% % 
% To split the supplementary pages from the main paper, you can use \href{https://support.apple.com/en-ca/guide/preview/prvw11793/mac#:~:text=Delete%20a%20page%20from%20a,or%20choose%20Edit%20%3E%20Delete).}{Preview (on macOS)}, \href{https://www.adobe.com/acrobat/how-to/delete-pages-from-pdf.html#:~:text=Choose%20%E2%80%9CTools%E2%80%9D%20%3E%20%E2%80%9COrganize,or%20pages%20from%20the%20file.}{Adobe Acrobat} (on all OSs), as well as \href{https://superuser.com/questions/517986/is-it-possible-to-delete-some-pages-of-a-pdf-document}{command line tools}.

\end{document}